\pdfoutput=1

\documentclass[10pt,twocolumn,letterpaper]{article}
\usepackage{titling}
\usepackage{iccv}
\usepackage{times}
\usepackage{epsfig}
\usepackage{graphicx}
\usepackage{amsmath}
\usepackage{amssymb}

\usepackage{array}
\usepackage{multirow}
\usepackage{placeins}
\usepackage[normalem]{ulem}
\newcommand{\RNum}[1]{\uppercase\expandafter{\romannumeral #1\relax}}
\usepackage[title]{appendix}
\usepackage{authblk}
\usepackage{capt-of}
\usepackage{threeparttable}
\usepackage{diagbox}
\usepackage{booktabs}
\usepackage{threeparttable}
\usepackage{adjustbox}
\usepackage{multicol}
\usepackage{flushend,cuted}

\usepackage{color}
\usepackage[accsupp]{axessibility}

\usepackage{pdfpages}

\newcommand\blfootnote[1]{%
\begingroup 
\renewcommand\thefootnote{}\footnote{#1}%
\addtocounter{footnote}{-1}%
\endgroup 
}

\usepackage[pagebackref=true,breaklinks=true,letterpaper=true,colorlinks,bookmarks=false]{hyperref}

\iccvfinalcopy 

\def\iccvPaperID{3552} 

\ificcvfinal\fi

\begin{document}

\title{Multi-scale Matching Networks for Semantic Correspondence}

\setlength{\affilsep}{0.4em}


\author[1,2]{Dongyang Zhao}
\author[1]{Ziyang Song}
\author[1]{Zhenghao Ji}
\author[3]{Gangming Zhao}
\author[1,2]{Weifeng Ge $^*$}
\author[3]{Yizhou Yu}

\affil[1]{Nebula AI Group, School of Computer Science, Fudan University} 
\affil[2]{Shanghai Key Lab of Intelligent Information Processing} 
\affil[3]{Department of Computer Science, The University of Hong Kong}

\twocolumn[{
\maketitle 

\renewcommand\twocolumn[1][]{#1}%
\begin{center}
\centering
\vspace{-30pt}
\includegraphics[width=\textwidth]{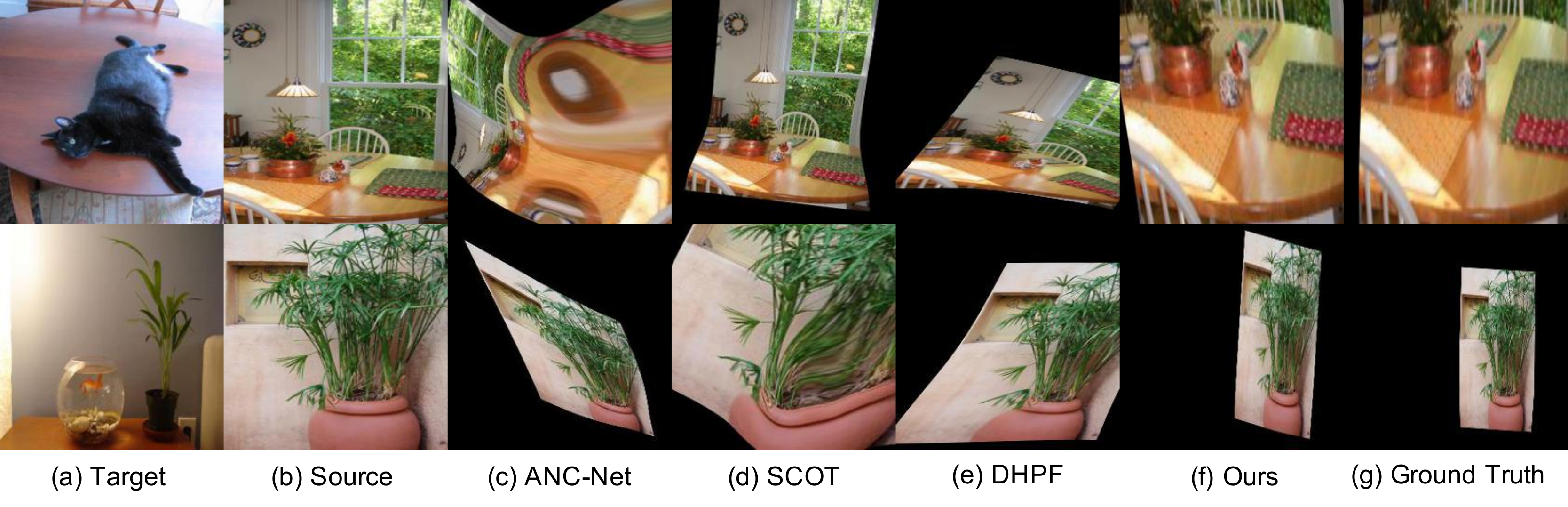}
\captionof{figure}{Dense correspondence prediction produced by state-of-the-art algorithms, including ANC-Net~\cite{li2020correspondence}, SCOT~\cite{liu2020semantic}, DHPF~\cite{min2020learning} and our multi-scale matching network. With the predicted key point pairs, images are warped with thin-plate splines algorithm~\cite{bookstein1989principal}.}\label{Fig:warping}
\end{center}

}]

\ificcvfinal\thispagestyle{empty}\fi
\blfootnote{*Corresponding author: wfge@fudan.edu.cn}

\begin{abstract}

 Deep features have been proven powerful in building accurate dense semantic correspondences in various previous works. However, the multi-scale and pyramidal hierarchy of convolutional neural networks has not been well studied to learn discriminative pixel-level features for semantic correspondence. In this paper, we propose a multi-scale matching network that is sensitive to tiny semantic differences between neighboring pixels. We follow the coarse-to-fine matching strategy and build a top-down feature and matching enhancement scheme that is coupled with the multi-scale hierarchy of deep convolutional neural networks. During feature enhancement, intra-scale enhancement fuses same-resolution feature maps from multiple layers together via local self-attention and cross-scale enhancement hallucinates higher-resolution feature maps along the top-down pathway. Besides, we learn complementary matching details at different scales thus the overall matching score is refined by features of different semantic levels gradually. Our multi-scale matching network can be trained end-to-end easily with few additional learnable parameters. Experimental results demonstrate that the proposed method achieves state-of-the-art performance on three popular benchmarks with high computational efficiency. The code has been released at \url{https://github.com/wintersun661/MMNet}.
\end{abstract}

\section{Introduction}
Finding pixel-wise correspondences between a pair of semantically similar images has been a longstanding fundamental problem in computer vision. They have been proven useful for many tasks including optical flow~\cite{hui2018liteflownet,sun2018pwc,sun2019models}, geometric matching~\cite{rocco2018end,melekhov2019dgc,truong2020glu}, disparity estimation~\cite{pang2017cascade,zhang2018activestereonet}, object recognition~\cite{duchenne2011graph,wohlhart2015learning,Zhang_2020_CVPR}, semantic segmentation~\cite{hur2016joint,larsson2019cross} and etc. Due to large intra-class variations in color, scale, orientation, illumination and non-rigid deformations, the problem of semantic correspondence remains very challenging.  With the breakthrough in representation learning, semantic correspondence has achieved impressive improvements in various scenarios.
 
Despite that deep features have improved matching accuracy significantly, the multi-scale and hierarchical structures of deep convolutional neural networks have not been explored thoroughly for semantic correspondence. It is well-known that convolutional neural networks can extract features of different semantic levels in a bottom-up manner~\cite{MD2014Visualizing}. Bottom convolutional layers close to the input image act like low level feature descriptors, and are sensitive to colors, edges, textures and other low level statistics. Top convolutional layers contain high level semantics which are similar among neighboring points in feature maps. Methods like NC-Net~\cite{Rocco18b}, DualRC-Net~\cite{li2020dual} and GOCor~\cite{GOCor_Truong_2020} use the features from the topmost layer as the feature representation. However, in semantic correspondence, the ambiguity between neighboring pixels in the topmost layer leads to inferior performance.  Hyperpixel flow~\cite{kong2016hypernet} and its extension~\cite{min2020learning} combine features at different semantic levels to generate reliable feature representation and achieve improved results. However, they have not thoroughly exploited the multi-scale and hierarchical structure of deep convolutional neural networks.

Given an image pair, human usually tends to glance at the whole images, and then compare details carefully to establish semantic correspondence. It is similar to a coarse-to-fine matching scheme. In a convolutional neural network, neurons in the top layers have larger receptive fields while neurons in the bottom layers have relatively small receptive fields, which means top layers are rich in semantics but have relatively weak localization ability while the bottom layers are strong in localization but have less semantics. It will be helpful to follow the multi-scale and hierarchical structure of convolutional neural networks to find semantic correspondence in a coarse-to-fine manner.

In this paper, we propose a new multi-scale matching network to produce reliable semantic correspondence by integrating features of different semantic levels hierarchically and learn complementary matching details in a coarse-to-fine manner. The multi-scale matching network consists of an encoder and a decoder. The encoder is a typical convolutional neural network pretrained on the ImageNet ILSVRC dataset~\cite{russakovsky2015imagenet}. It contains many layers to capture semantic information at different levels. We divide the feature maps in the encoder into five convolutional groups with respect to their resolutions. The decoder has two top-down hierarchical enhancement pathways across different scales. The first one is the feature enhancement pathway which upsamples spatially coarser, but semantically stronger feature maps and fuse them with features from lateral connections to hallucinate higher resolution features. The second one is the matching enhancement pathway that learns finer and complementary matching details to enhance coarser matching results from a lower resolution. We start from the first layer in the decoder to generate the coarsest matching results, and upsample and enhance them with complementary matching details at different semantic levels.     

To increase fine-grained details in feature maps, during intra-scale feature enhancement, we fuse all feature maps from the same convolutional group in the encoder not just the feature map of the last layer. We also design a transformer with a local self-attention mechanism to enhance features that are discriminative among neighboring pixels. Besides, we supervise matching detail learning at different scales to make sure the network learns reliable semantic correspondences. Our multi-scale matching network adds relatively few learnable parameters with little extra computational cost, and can be trained in an end-to-end manner easily.

In summary, the main contributions of this work can be summarized as follows:

\begin{itemize}
	\item We propose a multi-scale matching network that utilizes the multi-scale and hierarchical structure of deep convolutional neural network to learn semantic correspondences in a coarse-to-fine manner. Two top-down pathways in the decoder are built to couple the backbone encoder. The feature enhancement pathway increases  the representation power of feature maps with intra-scale enhancement and cross-scale enhancement. The matching enhancement pathway learns matching details that are complementary to matching results from coarser levels.
	\item We design a novel intra-scale feature enhancement module that simultaneously fuses all the feature maps in each convolutional group and further increases the discriminative ability of the fused feature map with a local transformer.
	\item Experimental results demonstrate that our multi-scale matching network achieves state-of-the-art performance on multiple popular benchmarks, including PF-PASCAL~\cite{ham2017proposal}, CUB ~\cite{cub} and SPair-71k~\cite{spair}.

\end{itemize}

\begin{figure*}[t]
	\centering
	\includegraphics[width=1.0\linewidth]{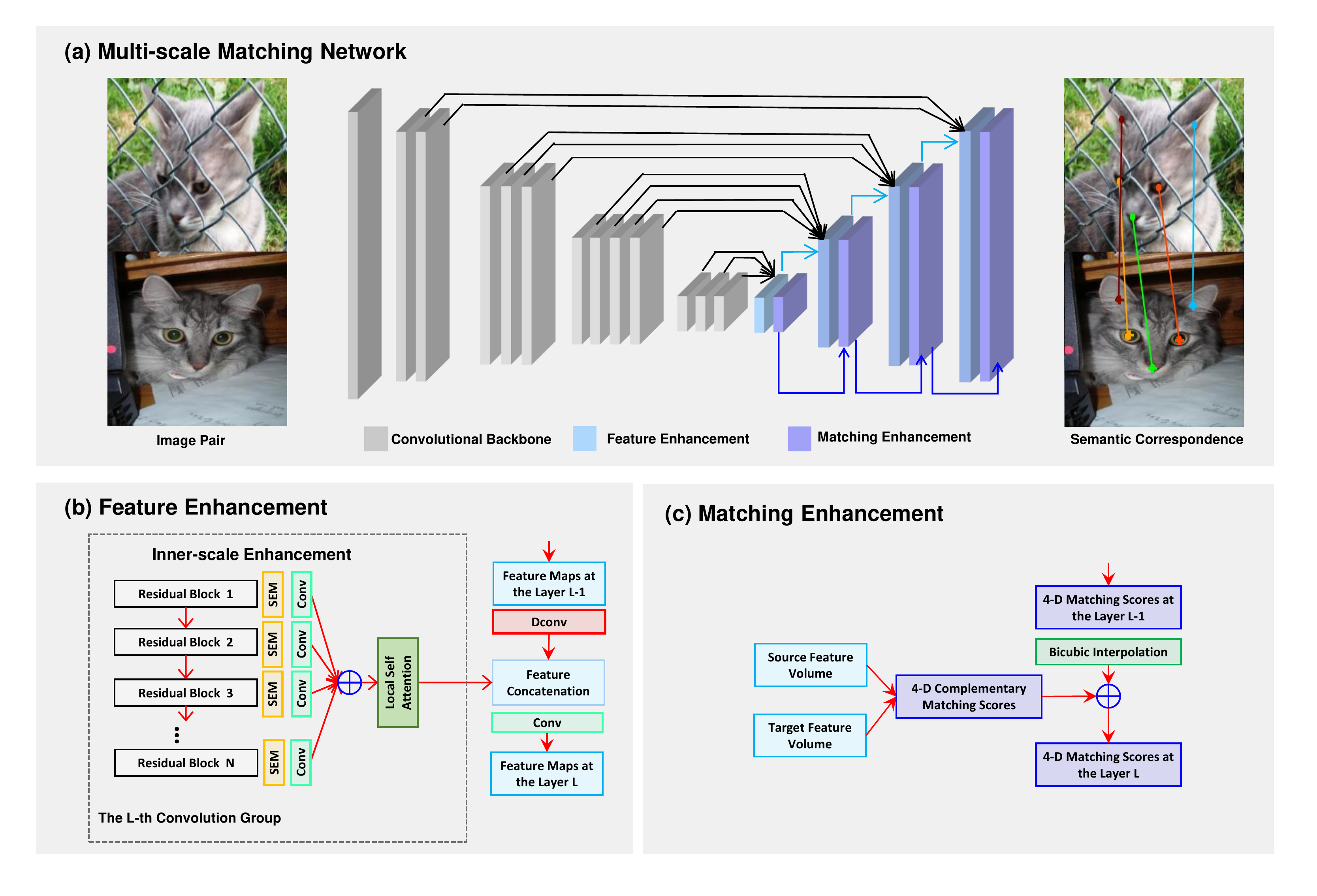}
	\caption{Illustration of the multi-scale matching network. Our multi-scale matching network contains a convolutional backbone, and a top-down feature and matching enhancement pathway. In feature enhancement, coarser but semantically richer features are upscaled and combined with finer but semantically weaker features to enhance the discrminaitve ability of pixel-level features.  Then the matching enhancement module directly upsamples the matching results of the previous scale and adds it with the current matching details to learn complementary correspondences across scales.}
	\label{Fig:mmn}
\end{figure*}

\section{Related Work} \label{related}
\noindent\textbf{Semantic Correspondence.} Methods for semantic correspondence can be roughly categorized into several groups: handcrafted feature based methods~\cite{lowe2004distinctive, bay2006surf, rublee2011orb, dalal2005histograms,tola2009daisy}, learnable feature based methods~\cite{li2020correspondence, li2020dual, min2020learning,GOCor_Truong_2020}, graph matching and optimization based methods~\cite{wang2019learning,liu2020semantic,Zhang_2020_CVPR,8661507}, methods focusing on geometry displacement~\cite{Cho_2015_CVPR,kanazawa2016warpnet,han2017scnet,ham2017proposal,truong2020glu}, and etc. Hand crafted features, such SIFT~\cite{lowe2004distinctive}, HOG~\cite{taniai2016joint} and DAISY~\cite{tola2009daisy}, design robust feature descriptors with low level statistics. In NC-Net~\cite{Rocco18b}, DualRC-Net~\cite{li2020dual} and GOCor~\cite{GOCor_Truong_2020}, high level semantic features of convolutional neural networks are used to build dense correspondences beween image pairs. SCOT~\cite{liu2020semantic} and DeepEMD~\cite{Zhang_2020_CVPR} formulate the semantic correspondence as an optimal transport problem and give closed-form solutions. PCA-GM~\cite{wang2019learning} and other graph matching based methods focus on solving a general quadratic assignment programming (QAP) problem to get matching results. Besides, PHM~\cite{Cho_2015_CVPR,ham2017proposal} and SCNet~\cite{han2017scnet} develop the probabilistic Hough matching in a Bayesian probability framework to model the geometry displacement of objects between two images.

In this paper, we learn semantic correspondence in a coarse-to-fine manner with a top-down matching enhancement scheme. By such hierarchical matching scheme, semantics at different levels and scales are fused and enhanced to get accurate pixel-wise correspondences.  

\noindent\textbf{Multi-scale Feature Fusion.} Multi-scale feature fusion can improve the representation ability of features in many tasks, including object detection~\cite{lin2017feature,zhao2019m2det}, semantic segmentation~\cite{ronneberger2015u,liu2018path} and semantic correspondence~\cite{min2019hyperpixel,min2020learning}. Feature pyramid networks (FPN~\cite{lin2017feature}) build a decoder with a top-down pathway and lateral connections, and achieve impressive results on object detection. Hyperpxiel flow (HPF~\cite{min2019hyperpixel}) searches the most informative convolutional feature to get superior results on semantic correspondence. Its extension Hypercolumns~\cite{min2020learning} designs a learning algorithm to select convolutional features in different layers in a much more efficient learning scheme.

Different from FPN~\cite{lin2017feature}, in the top-down pathway of our multi-scale matching network, feature maps in every layer of a convolutional group are fused to generate the lateral connection, not just the output of the last layer. Compared with HPF~\cite{min2019hyperpixel} and Hypercolumns~\cite{min2020learning}, our multi-scale learning scheme is much more flexible with simple top-down and lateral connections, and can benefit from the multi-scale and pyramid hierarchy more efficiently.. 

\noindent\textbf{4-D Correlation.} The 4-D correlation between two feature volumes is popular in semantic correspondence learning which calculates the matching scores densely. NC-Net~\cite{Rocco18b} analyzes neighborhood consensus patterns in the 4-D correlation space to get reliable dense correspondences. ANC-Net~\cite{li2020correspondence} introduces a set of non-isotropic 4-D convolution layers to capture adaptive neighborhood consensus. In this paper, we don't normalize the feature maps with $L_2$ normalization as that in NC-Net~\cite{Rocco18b} and ANC-Net~\cite{li2020correspondence}. We simply use the 4-D correlation tensor of two feature maps which stores the pair-wise scalar products as the matching score tensor. Then this 4-D correlation tensor is normalized with softmax to get the matching probability of every feature point. Experiments demonstrate that our proposed method without $L_2$ normalization can achieve impressive results. 

\noindent\textbf{Transformer.} Transformers have led to a series of breakthroughs in computer vision~\cite{zhu2020deformable,carion2020end,zhang2020feature} and natural language processing~\cite{vaswani2017attention, 2018BERT, 2019Fine}. In~\cite{vaswani2017attention}, elements in a sequence are encoded with a self-attention mechanism uniformly. While, in our local self attention, only neighborhood pixels are considered to enhancement the local patterns. Experimental results show that this local property in our local self attention works quite well for semantic correspondence.

\begin{figure*}[t]
	\centering
	\includegraphics[width=1.0\linewidth]{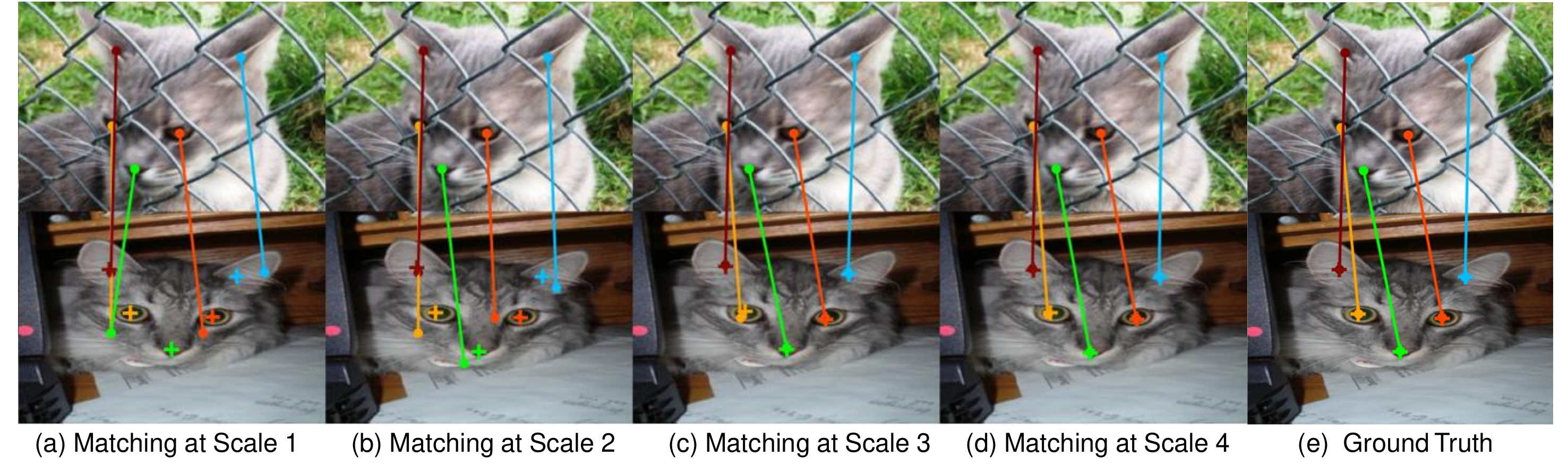}
	\caption{Matching results at different feature resolutions. From left to right, displacements between predictions and the destination points are reduced with the increase of the feature resolution.}
	\label{Fig:4res_dif}
\end{figure*}

\section{Multi-scale Matching Networks} \label{method}
Multi-scale matching networks utilize the multi-scale and hierarchical structures of deep convolutional neural networks to get discriminative pixel-level semantics for semantic correspondence. Figure~\ref{Fig:mmn} gives an overview of the proposed method. Given a pair of images $\left ( \boldsymbol{I}^s, \boldsymbol{I}^t \right )$ and the ground truth of their matched key points ${\mathcal{M}}_{gt} = \left \{ m_i = \left ( \boldsymbol{p}_i^s , \boldsymbol{p}_i^t \right ) |  i=1,...,K \right \}$, our multi-scale matching network adopts a top-down feature enhancement scheme and a coarse-to-fine matching enhancement scheme respectively. 
In the feature enhancement, we design a local self attention which models dependencies between neighborhood neurons to reduce the semantic ambiguity. In the matching enhancement, we learn the matching details that are complementary at different semantic levels. Besides, different from that in~\cite{Rocco18b,li2020correspondence}, feature volumes of two images are multiplied together directly without $L_2$ normalization to get 4-D matching scores. This scheme ensures the pixel-level similarity can be learnt with deep neural networks efficiently. Then we can train the multi-scale matching network by adding supervision at different scales in an end-to-end manner.

\paragraph{Network Structure.} A typical convolutional neural network for image classification~\cite{krizhevsky2012imagenet,he2016deep} usually has five convolutional groups with different resolutions. Here we adopt ResNet~\cite{he2016deep} as our backbone and follow the feature pyramid network (FPN~\cite{lin2017feature}) to broadcast semantics from high level layers to low level layers with a top-down pathway and lateral connections as shown in Figure~\ref{Fig:mmn} (a). Then in the decoding part, we get 4 feature maps with successive increasing resolutions by a scaling factor 2. 

\paragraph{Feature Enhancement.} There are feature enhancements both in the same scale and across different scales. During intra-scale enhancement, unlike FPN~\cite{lin2017feature} which only takes the feature map of the last residual block in each convolutional group, we simultaneously fuse the feature maps of all residual blocks in the same group to capture semantics at different levels. To enlarge the receptive field of a single neuron and capture semantics at different scales, every feature map passes a scale enhancement module (SEM~\cite{he2019bi}) before fusion. Then these feature maps are simply added together and go into a local self attention module to finish intra-scale enhancement. 

The local self attention module is designed to specialize every feature point. As common transformers~\cite{vaswani2017attention,zhang2020feature}, our local self attention operates on {\em{queries}}(Q), {\em{keys}}(K) and {\em{values}}(V) with an input feature map $\boldsymbol{X} \in \mathbb{R}^{C \times H \times W}$, and output a transformed version $\widetilde{\boldsymbol{X}}$ with the same shape as $\boldsymbol{X}$. For every location in $\boldsymbol{X}$, we select its $r \times r$ neighborhood to conduct the self-attention. Feature vectors of the neighborhood locations in $\boldsymbol{X}$ are collected, and then we get a neighborhood feature tensor $\boldsymbol{X}^{\prime} \in \mathbb{R}^{C \times H \times W \times r \times r}$. We send the input feature map $\boldsymbol{X}$ into the {\em{query}} transformation function $\mathcal{F}_q$, and send the neighborhood feature tensor $\boldsymbol{X}^{\prime}$ into the {\em{key}} and {\em{value}} transformation functions $\mathcal{F}_k$/$\mathcal{F}_v$ respectively. The {\em{query}}/{\em{key}}/{\em{value}} transformation functions are implemented with $1 \times 1$ convolutions followed by ReLU activations respectively.
The local interaction of our transformer is written as,
\begin{equation}
	\begin{aligned}
		\widetilde{\boldsymbol{X}}_i  = \boldsymbol{X}_i + \mathcal{G}\left ( \mathcal{F}_v\left ( \boldsymbol{X}^{\prime}_i \right ) \delta \left ( \mathcal{F}_q\left ( \boldsymbol{X}_i \right ) ^ \mathrm{ T }\mathcal{F}_k\left ( \boldsymbol{X}^{\prime}_i \right )  \right )  ^ \mathrm{ T } \right ),
	\end{aligned}
	\label{eq:localtrans}
\end{equation}
where $\boldsymbol{X}_i$ is the feature vector of the $i^{th}$ grid cell in $\boldsymbol{X}$, $\boldsymbol{X}^{\prime}_i$ is the neighborhood feature tensor of the $i^{th}$ grid cell,  $\mathcal{F}_q\left ( \boldsymbol{X}_i \right ) \in \mathbb{R}^{C^{\prime}}$ is $i^{th}$ {\em{query}}, $\mathcal{F}_k\left ( \boldsymbol{X}^{\prime}_i \right ) \in \mathbb{R}^{C^{\prime} \times r^2}$ and $\mathcal{F}_v\left ( \boldsymbol{X}^{\prime}_i \right ) \in \mathbb{R}^{C^{\prime} \times r^2}$ is $i^{th}$ {\em{key}}/{\em{value}} pair, $\delta$ is the SoftMax operation, and  $\mathcal{G}$ is a transformation function implemented with $1\times 1$ convolution.

After local self feature enhancement, we perform cross scale feature enhancement. Like the top-down pathway in FPN~\cite{lin2017feature}, the feature map from the previous matching stage are upsampled with a deconvolutional layer and are concatenated with $\widetilde{\boldsymbol{X}}$ before going through another convolutional layer. Then the enhanced feature map can be used to calculate the matching scores of an image pair and enhance features in the next stage. 

\paragraph{Matching Enhancement.} Unlike many cascaded methods such as FPN~\cite{lin2017feature} and BDCN~\cite{he2019bi} where results of different scales are jointly fused, we enhance the matching result in a top-down manner by learning matching complement at different scales. For a pair of images $\left ( \boldsymbol{I}^s, \boldsymbol{I}^t \right )$, denote their feature maps at the $l^{th}$ scale with $\boldsymbol{X}^{s}_l$ and $\boldsymbol{X}^{t}_l$ respectively. $\boldsymbol{X}^{s}_l$ and has the resolution $H^s_l \times W^s_l$, and $\boldsymbol{X}^{t}_l$ has the resolution $H^t_l \times W^t_l$. We calculate the exhaustive pairwise products between $\boldsymbol{X}^{s}_l$ and $\boldsymbol{X}^{t}_l$, and store the results in a 4-D tensor $\widetilde{\boldsymbol{S}}_{l} \in \mathbb{R}^{H^s_l \times W^s_l \times H^t_l \times W^t_l}$ referred to as the matching score. $\boldsymbol{S}_l\left (  i,j,m,n \right )$ is the match score between the $\left (  i,j \right )$ grid cell in $\boldsymbol{X}^{s}_l$ and the $\left (  m,n \right )$ grid cell in $\boldsymbol{X}^{t}_l$. Given the matching score $\boldsymbol{S}_{l+1} \in \mathbb{R}^{H^s_{l+1} \times W^s_{l+1} \times H^t_{l+1} \times W^t_{l+1}}$ from the previous scale, the matching complementation is conducted as follows,
\begin{equation}
	\begin{aligned}
		\boldsymbol{S}_{l} = \widetilde{\boldsymbol{S}}_{l} + \mathcal{U}\left ( \boldsymbol{S}_{l+1} \right ) ,
	\end{aligned}
	\label{eq:matchcom}
\end{equation}
where $\mathcal{U}$ is the 4-D bicubic upscaling interpolation. Then the $l$-th scale just learns the matching residuals that are complementary with the matching results in the $(l+1)$-th scale. Note that we start from the $5$-th scale with the highest semantic level where $\boldsymbol{S}_{5} = \widetilde{\boldsymbol{S}}_{5}$, and end at the $2$-nd scale. Figure~\ref{Fig:4res_dif} visualizes the improvements caused by the matching enhancement. Given these four matching results, we test their performance on the validation set, and select the scale that has the best performance to conduct testing.

\paragraph{Correspondence Learning with Rich Supervision.}  Labeling dense semantic correspondences of image pairs requires huge amount of human labours which is impractical in real applications. We evaluate the effectiveness of the proposed method on existing datasets with sparse key-point annotations including PF-PASCAL~\cite{ham2017proposal}, CUB ~\cite{cub} and SPair-71k~\cite{spair}. The sparse key-point annotation stores many one-tn-one mapping between images, where each mapping can be viewed as a probability distribution of a pixel in a source image $\boldsymbol{I}^s$ matched with all pixels in a target image $\boldsymbol{I}^t$. Then these annotations can be utilized in a straightforward way to train a CNN model for semantic matching by minimizing the distance between the matching distribution and the ground-truth distribution.  To benefit from the multi-scale matching mechanism further, we design loss functions at every scale to supervise the learning process. This rich supervision style leads to much more accurate matching results as stated in Table \ref{Tab:ablation_data}.

For a key point $\boldsymbol{p}_i^s$ in the source image, the matching score $\boldsymbol{S}_{l}(\boldsymbol{p}_i^s) \in \mathbb{R}^{H^t_l \times W^t_l}$ with the target image in the $l$-th scale is denoted in a 2-D form. Since deep features have very strong discriminative ability, we simply get the matching probability matrix $\boldsymbol{P}_{l}(\boldsymbol{p}_i^s)$ by applying the SoftMax operation spatially. We first rescale the key-points in ${\mathcal{M}}_{gt}$ to the same resolution as the feature maps at different scales. Then following ANC-Net~\cite{li2020correspondence}, we pick its four nearest neighbours and set their probability according to distance to establish the 2-D ground-truth matching probabilities at every scale. Then we apply 2-D Gaussian smoothing of size 3 on that probability map. Our training objectives for semantic matching is then,
\begin{equation}
	\begin{aligned}
		\mathcal{L} = \sum _{l} \alpha _ l \left [ \mathcal{B} \left ( {\boldsymbol{P}}_{l}(\boldsymbol{p}_i^s), \widetilde{\boldsymbol{P}}_{l}(\boldsymbol{p}_i^s) \right ) + \mathcal{B} \left ( {\boldsymbol{P}}_{l}(\boldsymbol{p}_i^t), \widetilde{\boldsymbol{P}}_{l}(\boldsymbol{p}_i^t) \right ) \right ],
	\end{aligned}
	\label{eq:matchloss}
\end{equation} 
where $\alpha_l$ (=1) is the weight at the $l$ scale, $\mathcal{B}$ is the binary cross entropy loss, and $\widetilde{\boldsymbol{P}}_{l}(\boldsymbol{p}_i^s) $ and $\widetilde{\boldsymbol{P}}_{l}(\boldsymbol{p}_i^t) $ are the ground-truth probability map of the key-point pair $\left ( \boldsymbol{p}_i^s , \boldsymbol{p}_i^t \right )$.

\section{Experiments}
\begin{table}
\centering
\normalsize
\begin{adjustbox}{max width=80 mm}
\begin{tabular}{l| c c c| c|c }
\toprule[1pt]
     \multirow{2}*{Methods}&\multicolumn{3}{c|}{PF-PASCAL}&CUB&time\\
     & 0.05&0.1&0.15& 0.1& (ms)\\\hline
     PF$_\text{HOG}$ \cite{ham2017proposal}  & 31.4 & 62.5 & 79.5&-&- \\ 
     CNNGeo$_\text{ResNet-101}$ \cite{cnngeo}  & 41.0 & 69.5 & 80.4&-&- \\ 
     A2Net$_\text{ResNet-101}$ \cite{a2net}  & 42.8 & 70.8 & 83.3& - &- \\ 
     SFNet$_\text{ResNet-101}$ \cite{sfnet}  & 53.6 & 81.9 & 90.6& - &- \\ 
     DCTM$_\text{CAT-FCSS}$ \cite{dctm}  & 34.2 & 69.6 & 80.2&-&-   \\ 
     WeakAlign$_\text{ResNet-101}$ \cite{rocco2018end}  & 49.0 & 74.8 & 84.0&-&-  \\ 
    SCNet$_\text{VGG-16}$ \cite{han2017scnet}  & 36.2 & 72.2 & 82.0&- &- \\ 
    RTNs$_\text{ResNet-101}$ \cite{rtns} & 55.2 & 75.9 & 85.2&-&-\\
    UCN$_\text{GoogLeNet}$ \cite{ucn} & -&55.6&-&48.3&-\\
    UCN$_\text{ResNet-101}$ \cite{ucn} & -&75.1&-&52.1&-\\
    NC-Net$_\text{ResNet-101}$ \cite{Rocco18b}  & 54.3 & 78.9 & 86.0& 64.7&393\\
    DCCNet$_\text{ResNet-101}$ \cite{dccnet}  & 55.6 & 82.3 & 90.5&66.1 &-\\
    HPF$_\text{ResNet-50}$ \cite{min2019hyperpixel}   & 60.5 & 83.4& 92.1&- &-\\
    HPF$_\text{ResNet-101}$ \cite{min2019hyperpixel}   & 60.1 & 84.8& 92.7& - &-\\
    HPF$_\text{ResNet-101-FCN}$ \cite{min2019hyperpixel}   & 63.5& 88.3& 95.4&-  &- \\
    DHPF$_\text{ResNet-50}$ \cite{min2020learning}   & 72.6 & \textbf{88.9} & \textbf{94.3}&-  &55\\
    DHPF$_\text{ResNet-101}$ \cite{min2020learning}   & 75.7 & \textbf{90.7}& \textbf{95.0}&- &95\\
    SCOT$_\text{ResNet-101}$ \cite{liu2020semantic}   & 63.1 & 85.4& 92.7& - &180\\
    SCOT$_\text{ResNet-101-FCN}$ \cite{liu2020semantic}   & 67.3 & 88.8& 95.4&- &109 \\
    ANC-Net$_\text{ResNet-101}$ \cite{li2020correspondence}   & - & 83.7& -& 69.6 &600 \\
    ANC-Net$_\text{ResNeXt-101}$ \cite{li2020correspondence}   & - & 88.7& -&74.1 &- \\
    ANC-Net$_\text{ResNet-101-FCN}$ \cite{li2020correspondence}   & - & 86.1& -& 72.4 &- \\
    \hline
    MMNet$_\text{ResNet-50}$  & \textbf{75.3} & 88.0& 93.2&\textbf{80.6} &51\\
    MMNet$_\text{ResNet-101}$  & \textbf{77.6} & 89.1& 94.3&\textbf{81.8} &86\\
    MMNet$_\text{ResNeXt-101}$  & \textbf{78.9} & \textbf{90.3}& \textbf{94.4}&\textbf{83.1}& 101\\
    MMNet$_\text{ResNet-101-FCN}$  & \textbf{81.1} & \textbf{91.6}& \textbf{95.9}&\textbf{87.0} &87\\
    \bottomrule[1pt]
\end{tabular}
\end{adjustbox}
\vspace{2pt}
\caption{Comparison with state-of-the-art algorithms in PCK and speed on PF-PASCAL~\cite{ham2017proposal} and CUB ~\cite{cub} dataset. Subscripts of the method names indicates the backbone used.}\label{Tab:quant_results}
\end{table}

\begin{table*}
\centering
\normalsize
\begin{adjustbox}{max width=180mm}
\begin{threeparttable}


\begin{tabular}{l| c c c c c c c c c c c c c c c c c c|c}
\toprule[1pt]
    Methods & aero & bike & bird & boat & bottle & bus & car & cat & chair & cow & dog& horse& mbike& person& plant& sheep& train & tv & all\\ \hline
    CNNGeo \cite{cnngeo}  & 23.4 & 16.7 & 40.2& 14.3& 36.4& 27.7& 26.0& 32.7& 12.7& 27.4& 22.8& 13.7& 20.9& 21.0& 17.5& 10.2& 30.8& 34.1& 20.6\\ 
    A2Net \cite{a2net}  & 22.6 & 18.5& 42.0& 16.4& 37.9& 30.8& 26.5& 35.6& 13.3& 29.6& 24.3& 16.0& 21.6& 22.8& 20.5& 13.5& 31.4& 36.5&22.3 \\
    WeakAlign \cite{rocco2018end}  & 22.2 & 17.6& 41.9& 15.1& 38.1& 27.4& 27.2 & 31.8& 12.8& 26.8& 22.6& 14.2& 20.0& 22.2&17.9&10.4&32.2&35.1&20.9\\
    NC-Net \cite{Rocco18b}   & 17.9 & 12.2& 32.1& 11.7 & 29.0 & 19.9&16.1 &39.2 &9.9&23.9&18.8&15.7&17.4&15.0&14.8&9.6&24.2&31.1&20.1 \\
    HPF \cite{min2019hyperpixel}   & 25.3 & 18.5& 47.6& 14.6& 37.0& 22.9& 18.3& 51.1& 16.7& 31.5& 30.8& 19.1& 23.7& 23.8& 23.5&14.4&30.8&37.2&27.2\\
    HPF \cite{min2019hyperpixel}   & 25.2 & 18.9& 52.1& 15.7& 38.0& 22.8& 19.1& 52.9& 17.9& 33.0& 32.8& 20.6& 24.4& 27.9& 21.1& 1.9& 31.5& 35.6&28.2\\
    DHPF \cite{min2020learning}   & 38.4 & 23.8& \textbf{68.3}& 18.9& 42.6& 27.9& 20.1& 61.6& 22.0& 46.9 & 46.1& 33.5& 27.6& \textbf{40.1}& 27.6&28.1&49.5&46.5&37.3 \\
    SCOT \cite{liu2020semantic}   & 34.9 & 20.7& 63.8& 21.1& 43.5& 27.3& 21.3& \textbf{63.1}& 20.0& 42.9& 42.5& 31.1& 29.8& 35.0& 27.7& 24.4&48.4& 40.8& 35.6\\
    \hline
    MMNet  & 43.5 & 27.0& 62.4&27.3 & 40.1 & 50.1&37.5 & 60.0 & 21.0& 56.3& 50.3& 41.3&30.9&19.2&30.1&33.2&64.2&43.6&40.9 \\
    MMNet-FCN  & \textbf{55.9} & \textbf{37.0}& 65.0&\textbf{35.4} & \textbf{50.0} & \textbf{63.9}&\textbf{45.7} & 62.8 & \textbf{28.7}& \textbf{65.0}& \textbf{54.7}&\textbf{51.6}&\textbf{38.5}&34.6&\textbf{41.7}&\textbf{36.3}&\textbf{77.7}&\textbf{62.5}&\textbf{50.4} \\
    \bottomrule[1pt]
\end{tabular}

   \end{threeparttable}

\end{adjustbox}
\vspace{1pt}
\caption{Comparisons on SPair-71k~\cite{spair} with state-of-art methods. The backbone in methods listed is ResNet101~\cite{he2016deep}. The best results are reported in bold.}\label{Tab:quant_spair}
\end{table*}

\paragraph{Dataset.}
We conduct experiments on three popular benchmarks for semantic correspondence: PF-PASCAL~\cite{ham2017proposal}, CUB ~\cite{cub} and SPair-71k~\cite{spair}. The PF-PASCAL contains 1351 image pairs which are selected from all the 20 categories in PASCAL VOC \cite{pascal-voc}. We split the dataset as done in~\cite{han2017scnet} where approximately 700 image pairs are used for training, 300 image pairs are used for validation and 300 image pairs are used for test. The CUB dataset~\cite{cub} contains 11,788 images of 200 bird species with large intra-class variations. Each image is annotated with the locations of 15 key-parts. Following the protocol in~\cite{li2020correspondence}, we randomly sample 10,000 pairs from the CUB as training data and use the same test set provided by~\cite{cub-test}. SPair-71k is composed of total 70,958 image pairs in 18 categories with large view-point and scale variations. We use the same split proposed in~\cite{spair} where 53340, 5384, 12234 image pairs are used for training, validation and testing respectively. 

\paragraph{Evaluation metric.}Performances of different methods are evaluated using the percentage of correct key-points (PCK@$\alpha$). A point is considered \textit{correct} if the predicted point is within the circle of radius $\alpha \times d $ centering at the ground-truth point, where d is the longer side of an image or an object bounding box as in \cite{han2017scnet,Rocco18b,li2020correspondence,liu2020semantic,min2020learning}.

\paragraph{Implementation Details}

For fair comparison with state-of-the art methods, we use four different backbones including ResNet-50~\cite{he2016deep}, ResNet-101~\cite{he2016deep}, ResNeXt-101~\cite{xie2017aggregated} and ResNet101-FCN~\cite{he2016deep}. All backbone networks are pretrained on Image-Net1k classification set~\cite{krizhevsky2012imagenet} and then fine-tuned for corrrespondence task. 

The multi-scale matching network structure is visualized in Figure~\ref{Fig:mmn}. We only introduce additional parameters in feature enhancement. As shown in Figure~\ref{Fig:mmn} (b), we have many SEMs  each of which is followed by a $1 \times 1$ convolutional layer. We upscale the low resolution feature with a $4 \times 4$ deconvolutional layer whose stride is 2 at different scales. Then the upsampled feature map is concatenated with the output of the intra-scale feature enhancement, and pass a $3 \times 3$ convolution layer. Note that the SEM in our MMNet is in the same settings as in BDCN~\cite{he2019bi}, and the output channel numbers of both the convolutional layers and the deconvolutional layers is set to 21 to save computation cost.

During the training, we adopt SGD with momentum as our optimizer. The learning rate is set to 0.0005 for initialization and is decreased by 10 times every 10000 iterations. Momentum and weight decay are set to 0.9 and 0.0002 respectively. Learning rate is decreased by 10 times every 10,000 iterations. The batch-size is set to 5 for all experiments. The training will converge within 10000, 32000 and 30000 iterations for PF-PASCAL~\cite{ham2017proposal}, CUB ~\cite{cub} and SPair-71k~\cite{spair} respectively. All experiments are implemented with PyTorch~\cite{pytorch}, and run on NVidia TITAN RTX GPUs.


\begin{figure}[t]
	\centering
	\includegraphics[width=1.0\linewidth]{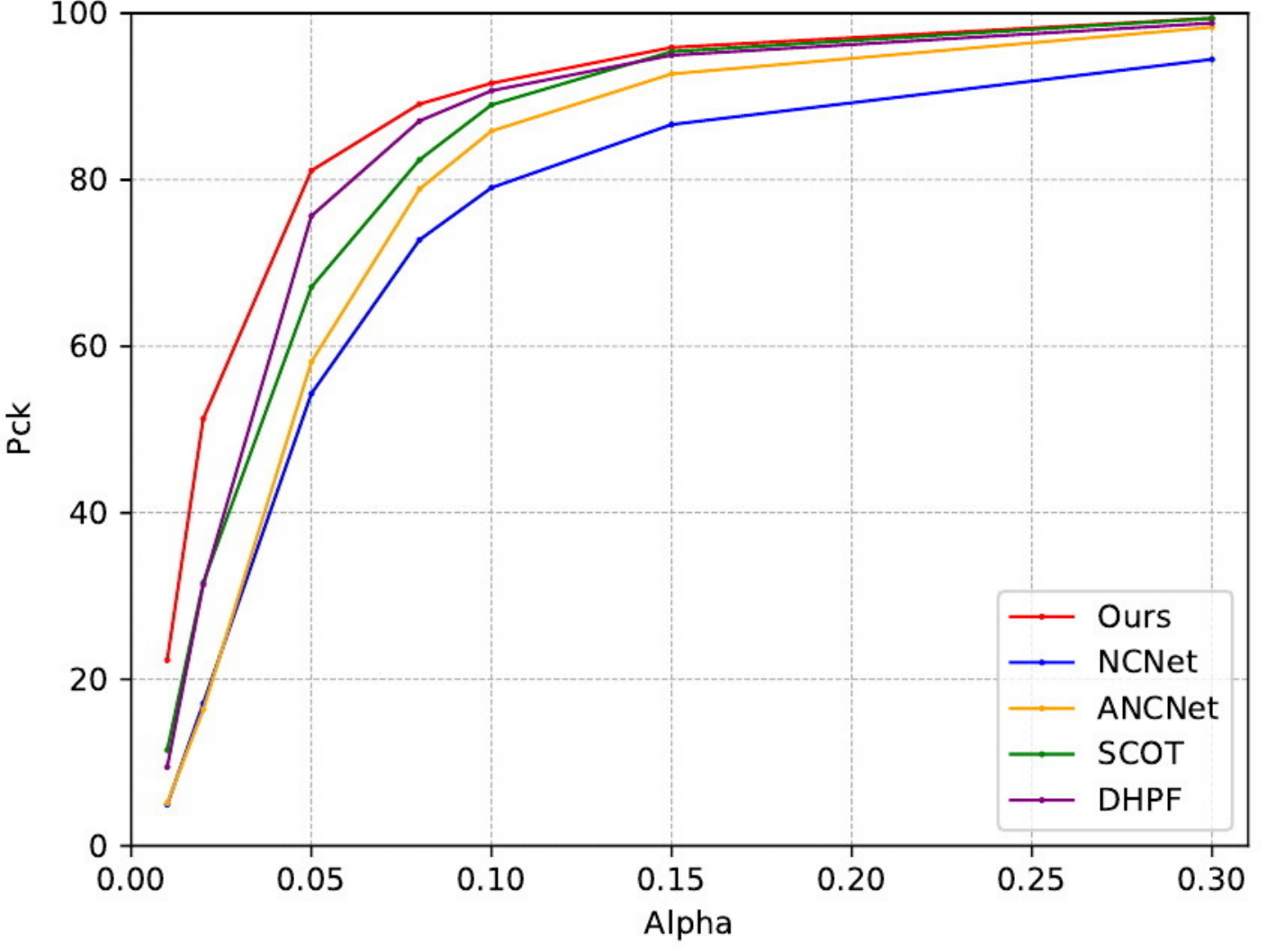}
	\caption{The PCK-$\alpha$ curves of our method and compared works on PF-PASCAL~\cite{ham2017proposal}.}
	\label{Fig:alpha}
\end{figure}

\subsection{Comparisons with State-of-the-art Methods}\label{comparisons}

For PF-PASCAL\cite{ham2017proposal}, our MMNet with ResNet101-FCN as the backbone outperforms all previous state-of-art methods with 81.1\% PCK@0.05, 91.6\% PCK@0.1, and 95.9\% PCK@0.15. When compared with ANC-Net~\cite{li2020correspondence} which also conducts end-to-end training with 4-D correlation, MMNet gets 5.4\%, 1.6\%, and 5.5\% increases on PCK@0.1 with three different backbones respectively. It demonstrates the effectiveness of the multi-scale feature learning and matching complementation. When compared with previous best SCOT~\cite{liu2020semantic} with ResNet101-FCN as the backbone, we achieve a significant improvement on PCK@0.05 by 13.8\%. We attribute this to that the end-to-end training of deep neural networks has a higher efficiency that optimization based methods to enforce one-on-one matching with discriminative features. We also compare with the multi-scale feature fusion based methods  HPF~\cite{min2019hyperpixel} and DHPF~\cite{min2020learning}. MMNet with ResNet101-FCN as the backbone outperforms HPF with the same backbone by 17.6\% PCK@0.05, 3.3\% PCK@0.1 and 0.5\% PCK@0.15. Since HPF doesn't conduct end-to-end training, it is reasonable that our MMNet gets better results. DHPF~\cite{min2020learning} selects features from the backbone, and get slightly better results on PCK@0.1 and PCK@0.15 with ResNet50, ResNet101 backbone. However, when $\alpha=0.05$, our MMNet get 1.9\% improvement on PCK. It may because MMNet are much more sensitive to small difference between neighborhood, and thus get better matching results with much more strict matching criteria. For CUB\cite{cub}, our MMNet outperforms all state-of-art algorithms with 87.0\% on PCK@0.1 and achieve prominent betters on three different backbone compared with ANCNet~\cite{li2020correspondence}.For SPair-71k\cite{spair}, our MMNet with ResNet101-FCN backbone outperforms the state-of-art algorithms by at least 13.1\% on PCK@0.1, which is a huge improvement. Among all listed methods in Table~\ref{Tab:quant_results}, our algorithm surpasses all state-of-art by a large margin on 15 of the 18 classes. This proves the effectiveness and robustness of our MMNet in establishing reliable matching. 

Figure~\ref{Fig:alpha} shows the results in comparison with state-of-art method with varying $\alpha$. When $\alpha$ is small, only points matched with the destination point closely is treated as a correct match, otherwise failure. When we increase $\alpha$, lager matching displacement will be allowed. It can be found that our MMNet achieves the best performance when $\alpha$ varies from 0.02 to 0.3. When $\alpha$ varies from 0.02 to 0.1, all algorithms will get improvements on PCK fast. It means the allowed match displacement influences the performance greatly. When $\alpha$ varies from 0.15 to 0.3, all methods get almost the same results with very high matching accuracy. This indicates too large $\alpha$ can not be used to measure the performance of different methods accurately. When $\alpha$ varies from 0.01 to 0.05, our MMNet outperforms other state-of-art methods by a clear margin all the time. It indicates the strong ability of MMNet in identifying neighborhood pixels. 

Besides, we also report the running speed to compare the computational efficiency of state-of-art methods. In Table~\ref{Tab:quant_results}, we get the comparable test speed with the previous best DHPF~\cite{min2020learning}, and are much faster than other methods.

\begin{figure*}[t]
	\centering
	\includegraphics[width=1.0\linewidth]{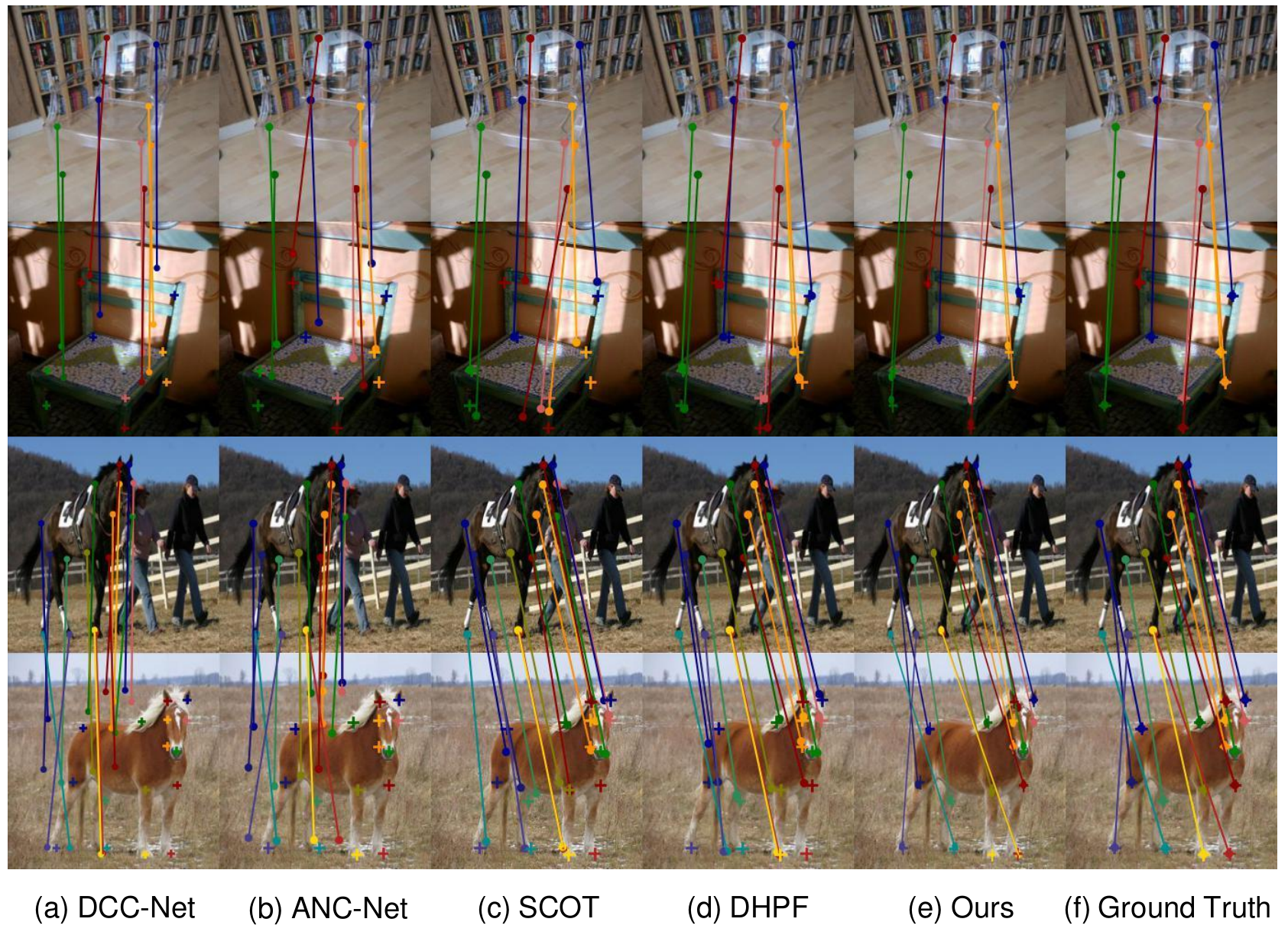}
	\caption{Visualization of the semantic correspondence. The odd rows are the source image, and the even rows are the target images. Destination key point are denoted with crosses. From left to right: (a) DCC-Net~\cite{dccnet}, (b) ANC-Net~\cite{li2020correspondence}, (c) SCOT~\cite{liu2020semantic}, (d) DHPF~\cite{min2020learning}, (e) ours MMNet and (f) the ground truth.}
	\label{Fig:msmc}
\end{figure*}

\subsection{Module Analysis with Ablations}\label{ablations}
To investigate the effectiveness of different modules, we conduct ablation study by replacing or removing a single component. All experiments are conducted on PF-PASCAL~\cite{ham2017proposal} with ResNet101-FCN as the backbone. PCKs are evaluated with $\alpha$ = 0.05, 0.1 and 0.15. Results are listed in Table~\ref{Tab:ablation_data}. First, we remove the local self attention, the performance drops by 1.2\% on PCK@0.05 which indicates that the contextual information around neighborhood pixels is very important. Then we remove the dense connections and only get the output of the last layer in every convolutional group, the performance of PCK@0.05 is 74.9\% which is 6.2\% lower than MMNet with dense connections. It shows fusing information at different layers in a convolutional group is greatly helpful. When we remove the cross-scale feature fusion, PCK@0.05, PCK@0.10 and PCK@0.15 decrease by 2.6\%, 1.9\% and 1.6\% respectively. If we don't conduct the complementary matching learning and get the matching score at every layer independently, PCK@0.05 drops by 1\%. At last, if we only supervise the learning at the last matching complementation layer with the highest resolution, the performance drops to 31.4\% on PCK@0.05 which is 49.7\% lower than the MMNet with rich supervision during training. It may indicate that supervision in multiple scales are very important to learn semantics at different levels.  


\begin{table}
\centering
\normalsize
\begin{tabular}{l |c c c}
\toprule[1pt]
     Methods & 0.05 & 0.1& 0.15 \\ \hline 
    MMNet  & 81.1 & 91.6 & 95.9 \\ \hline
    MMNet w/o local self attention   & 79.9 & 90.7& 95.1\\
    MMNet w/o dense connections   & 74.9 & 88.6 & 93.9\\
    MMNet w/o cross-scale fusion & 78.5 & 89.7& 94.3\\
    MMNet w/o complementation  & {80.0} & {89.3} & {94.3}\\ 
    MMNet w/o rich supervision   & 31.4 & 45.6 & 57.0\\ 
    MMNet w/o last layer & 77.9&90.9 &94.6 \\
    \bottomrule[1pt]
\end{tabular}
 \vspace{5pt}
\caption{Ablation results on various setting on PF-PASCAL~\cite{ham2017proposal} with ResNet101-FCN as backbone.}
\label{Tab:ablation_data}
\end{table}

\subsection{Qualitative Results and Visualization}\label{qualitatives}
We visualize the correspondence result by drawing the point-to-point matches and warping images with the predicted key point pairs respectively. In Figure~\ref{Fig:msmc}, the point-to-point matches are drawn by linking key point pairs with line segments. The ground truth matching is given at first as reference for visual comparison. It can be find that our MMNet matches all key points on horses correctly. Other state-of-the art methods, such as DCC-Net~\cite{dccnet}, ANC-Net~\cite{li2020correspondence}, SCOT~\cite{liu2020semantic} and DHPF~\cite{min2020learning}, will lead to large mismatch displacement or  many-to-one match. In Figure~\ref{Fig:warping}, images are warped based on matched key point pairs. For convenience, we warp the ground truth annotations as the reference. It can be found that our MMNet can matches the objects in images accurately. Especially in the first row, tables in the source and target images has very large viewpoint and appearance variations. But our MMNet can still match the corresponding key points accurately when other methods fail.

\section{Conclusion}
In this paper, we propose a multi-scale matching network that are coupled with the multi-scale and pyramidal hierarchy of deep convolutional neural networks to match semantic meaningful points in a coarse-to-fine manner. Our multi-scale matching network learn discriminative pixel-level semantics by a top-down feature and matching enhancement scheme. Thus discriminative features are identified and fused during the complementary matching learning process. To strengthen the representatve ability of individual pixels, we introduce a local self attention module to encode local contextual information to disambiguate the feature representation of neighborhood pixels. Extensive experiments on several popular benchmarks demonstrate that the proposed MMNet outperform existing state-of-the-art semantic correspondence algorithms. However, how to get reliable correspondence that can handle drastically changes in real applications still remains to be an open problem.

\FloatBarrier
{\small
\bibliographystyle{ieee_fullname}
\bibliography{experiments}
}

\clearpage
\makeatletter
\def\@thanks{}
\makeatother

\title{\emph{Supplementary Material} for \\
Multi-scale Matching Networks for Semantic Correspondence}

\makeatletter
\renewcommand{\@maketitle}{
\newpage
 \null
 \vskip 2em%
 \begin{center}%
  {\LARGE \@title \par}%
 \end{center}%
 \par} \makeatother

\renewcommand\thesection{\Alph{section}}
\setcounter{section}{0}
\setcounter{table}{0}
\setcounter{figure}{0}
\graphicspath{{figures/}}

    \onecolumn
    \maketitle
    \setlength{\parindent}{1em} 
    \setlength{\parskip}{0em}
    
    \section{Parameter Table}
 Images are resized to $224 \times 320$ for all datasets both in training and testing.
{\begin{table}[h]
   \centering
    
    \begin{adjustbox}{max width=165 mm}
    
    \begin{tabular}{l|c|p{3cm}<{\centering}|p{3cm}<{\centering}|p{3cm}<{\centering}|p{3cm}<{\centering}}
    \toprule[1pt]
    
      \multicolumn{2}{l|}{}& ResNet-50 & ResNet-101& ResNeXt-101& ResNet-101-FCN \\ \hline
     
     Conv Intra-fused& \multicolumn{4}{c}{conv: output channel = 21, kernel size = 1$\times$1 , stride = 1}\\  \hline
     Conv Cross-fused& \multicolumn{4}{c}{conv: output channel = 21, kernel size = 3$\times$3, stride = 1, padding = 1}\\  \hline
     Deconv& \multicolumn{4}{c}{transpose conv: output channel = 21, kernel size = 4$\times$4, stride = 2, padding = 1 , bias = False} \\ \hline
    \multirow{4}*{LSA}  &conv\_key&\multicolumn{3}{c}{conv: output channel = 10, kernel size = 1$\times$1} \\ &conv\_value&
     \multicolumn{3}{c}{conv: output channel = 10, kernel size = 1$\times$1}\\       
    & conv\_query &
     \multicolumn{3}{c}{conv: output channel = 10, kernel size = 1$\times$1}\\
     & conv\_aggregate &
     \multicolumn{3}{c}{conv: output channel = 21, kernel size = 1$\times$1} 
     \\ \hline \hline
     \multirow{3}*{\# Params} & backbone & $25.6\times10^{6}$ & $44.5.0\times10^{6}$ & $88.8\times10^{6}$ & $54.4\times10^{6}$ \\
     &MMNet w/o backbone &  $4.8\times10^{6}$ & $10.3\times10^{6}$ & $10.3\times10^{6}$ & $10.3\times10^{6}$ \\
     &MMNet &  $30.4\times10^{6}$ & $54.8\times10^{6}$ & $99.1\times10^{6}$ & $64.7\times10^{6}$ \\ \hline
     \multirow{3}*{FLOPs} & backbone & $11.8\times10^{9}$ & $22.4\times10^{9}$ & $47.0\times10^{9}$ & $127.6\times10^{9}$ \\
     &MMNet w/o backbone &  $3.1\times10^{9}$ &  $4.6\times10^{9}$ &  $4.6\times10^{9}$ &  $12.7\times10^{9}$ \\
     &MMNet &  $14.9\times10^{9}$ &  $27.0\times10^{9}$ & $51.6\times10^{9}$ &  $140.3\times10^{9}$\\

     \bottomrule[1pt]
    \end{tabular}
    \end{adjustbox}
    \vspace{10pt}
    \caption{Implementation details of MMNet, and numbers of parameters and FLOPs introduced by different modules. In this table, \emph{`Conv Intra-fused'} and \emph{`Conv Cross-fused'} denote the convolution operation in the intra-scale and cross-scale feature enhancements separately, \emph{`Deconv'} indicates the deconvolution operation to upscale the feature maps during the cross-scale feature enhancement, and \emph{`LSA'} is short for the local self attention module used at the end of the intra-scale feature enhancement.}
    \label{tab:my_label}
\end{table}
}

\paragraph{Analysis.}
MMNet introduces additional parameters only in its feature enhancement module. We follow BDCN~\cite{he2019bi} to set the parameters of scale enhancement modules. Each scale enhancement module contains four $3 \times 3$ convolution operations with dilation 1, 4, 8, 12 respectively. The output channel numbers of these convolution operations are set to 32.  The output channel numbers of other convolution and deconvolution operations are set to 21 to reduce the computational cost. We compare the additional parameters and FLOPs introduced by MMNet with four different backbones including ResNet-50, ResNet-101, ResNeXt-101 and ResNet-101-FCN. The additional parameters are no larger than 25\% of that of any backbone. For the FLOPs, we add at most 26.3\% computational cost when using ResNet-50 as the backbone. \\ 

\section{Additional Results}

\paragraph{Results with other backbones.}
We adapt our MMNet design with other backbones: VGG-16 and DeepLab-V3, the results are listed in \ref{tab:strong_backbone}. As can be seen, our model with ResNet-101-FCN backbone performs significantly best than others.
\begin{table}[] 
    \centering

    \begin{tabular}{l|c c c}
    \toprule[1pt]
    \multirow{2}*{Methods}&\multicolumn{3}{c}{PF-PASCAL}\\
    & 0.05&0.1&0.15\\ \hline
        MMNet$_\text{ResNet-101-FCN}$ & 81.1 & 91.6& 95.9\\
        MMNet$_\text{VGG-16}$ & 69.5 & 82.0& 88.4\\
        MMNet$_\text{DeepLabV3-ResNet101}$ & 73.3&85.3&92.3\\
        \bottomrule[1pt]
    \end{tabular}
\vspace{5pt}
    \caption{Experiments on different backbones. All experiments is conducted on PF-PASCAL.}
    \label{tab:strong_backbone}
\end{table}

\newpage
\begin{figure*}[h]
	\centering
	\includegraphics[width=0.58\linewidth]{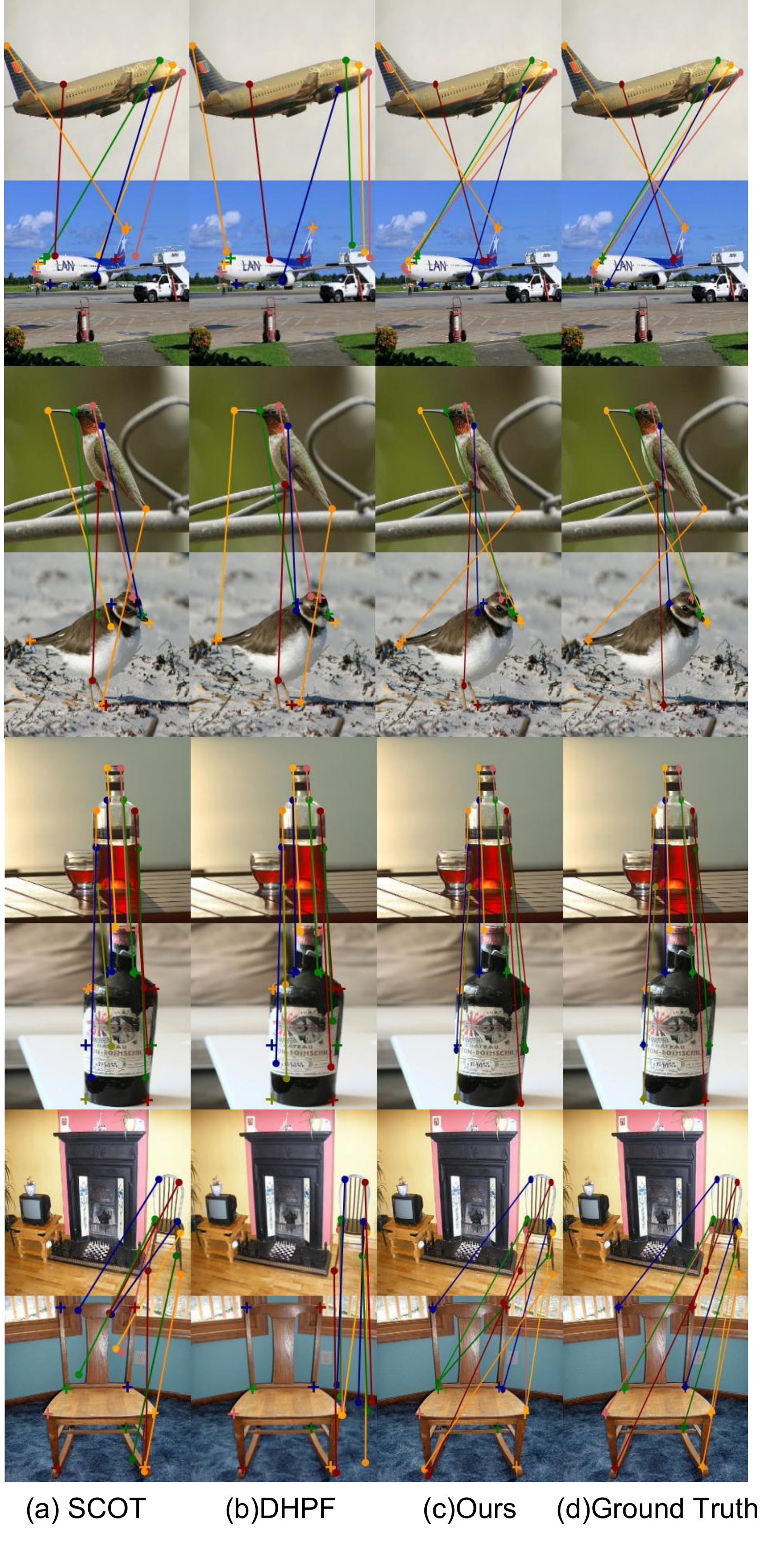}
	\caption{Key-point matching results on SPair-71k dataset \cite{spair} compared with SCOT \cite{liu2020semantic} and DHPF  \cite{min2020learning}. The odd rows are the source images,  and the even rows are the target images. Destination key points are denoted with crosses.}
	\label{Fig:kps_1}
\end{figure*}

\begin{figure*}[h]
	\centering
	\includegraphics[width=0.6\linewidth]{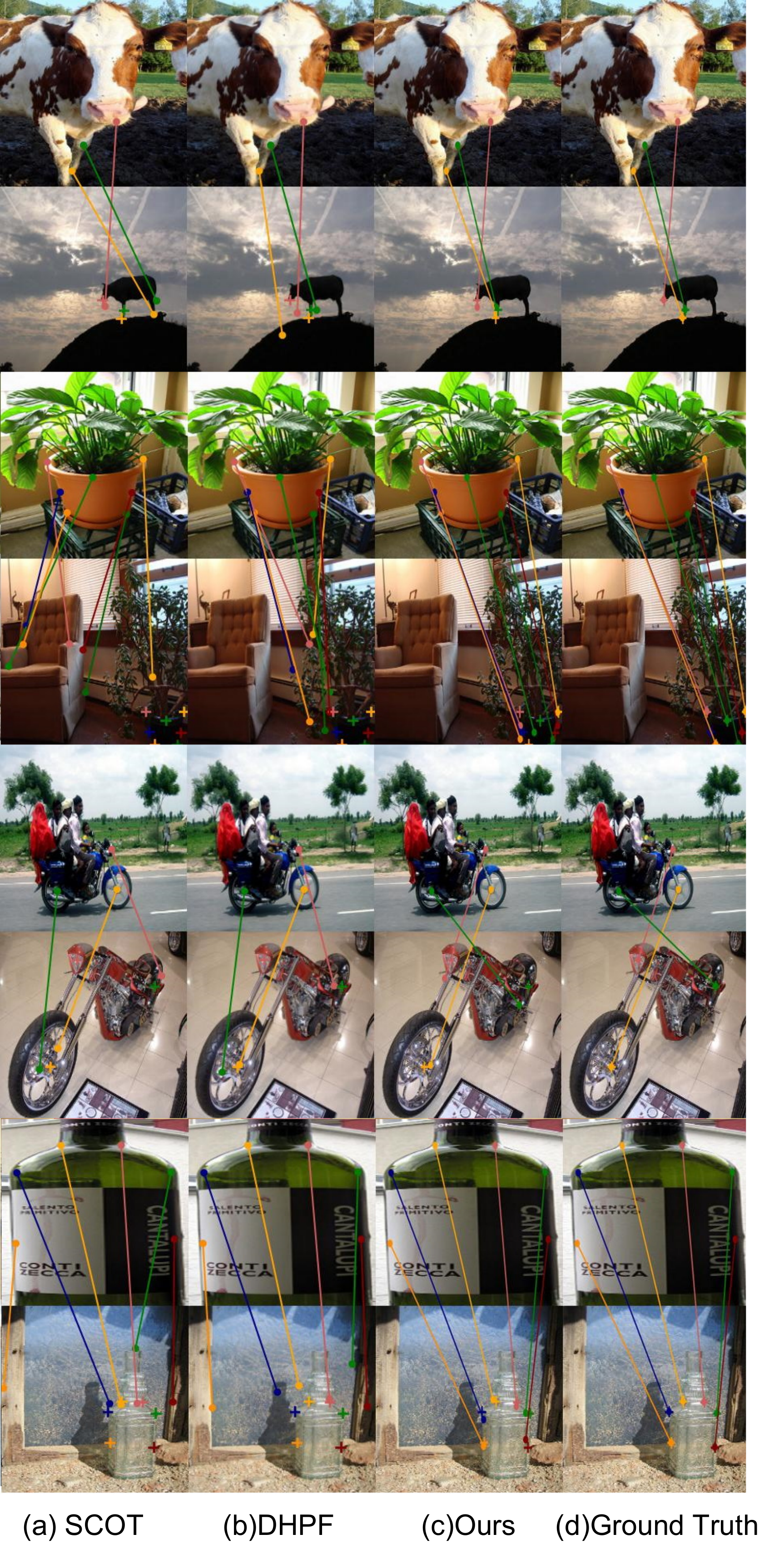}
	\caption{Key-point matching results on SPair-71k dataset  \cite{spair} compared with SCOT \cite{liu2020semantic} and DHPF \cite{min2020learning}. The odd rows are the source images,  and the even rows are the target images. Destination key points are denoted with crosses.}
	\label{Fig:kps_2}
\end{figure*}
\begin{figure*}[h]
	\centering
	\includegraphics[width=0.8\linewidth]{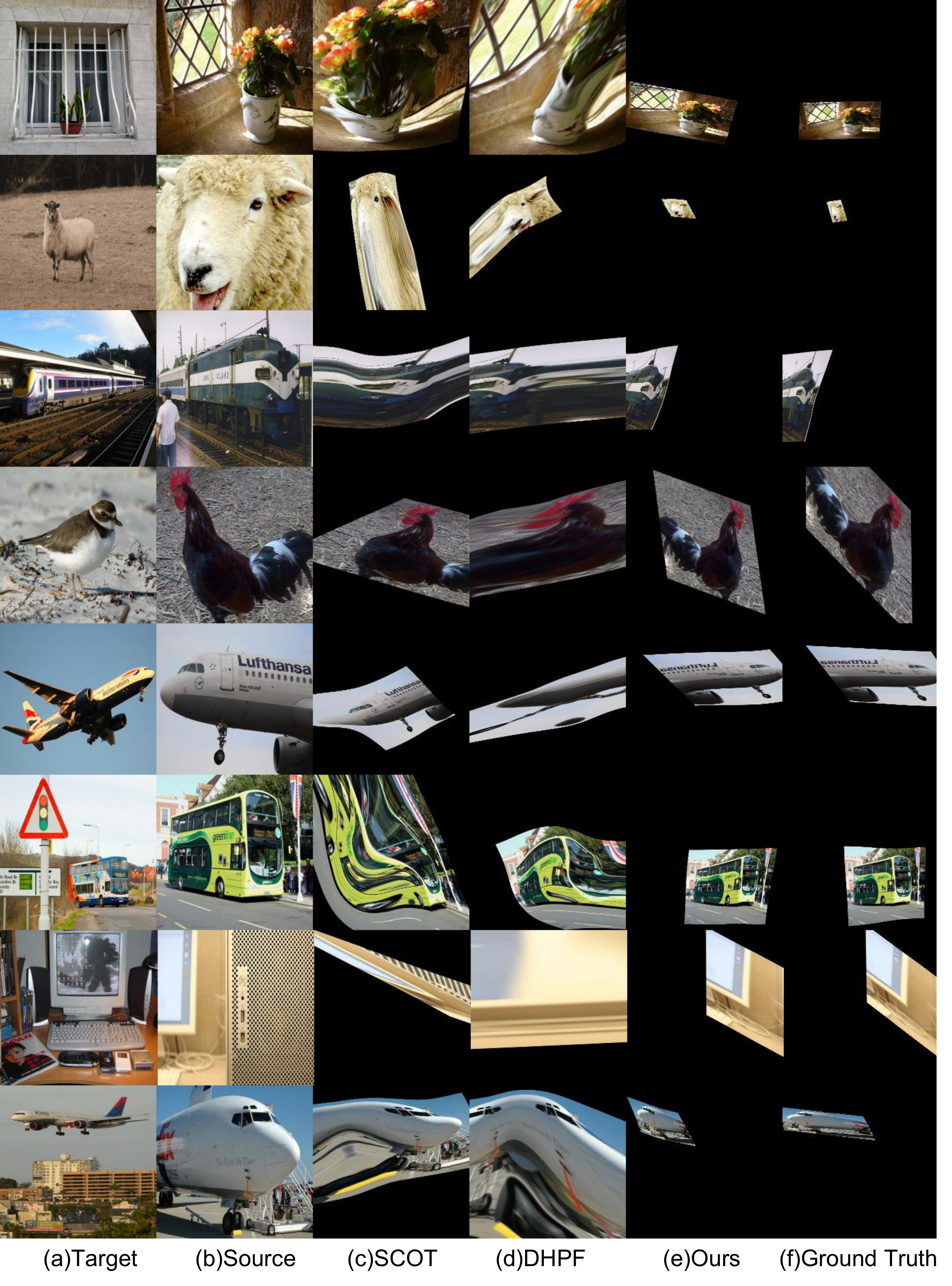}
	\caption{Warped images by thin-plate splines with the predicted key point pairs on SPair-71k dataset  \cite{spair} compared with SCOT \cite{liu2020semantic} and DHPF \cite{min2020learning}}
	\label{Fig:warp_1}
\end{figure*}
\begin{figure*}[h]
	\centering
	\includegraphics[width=0.8\linewidth]{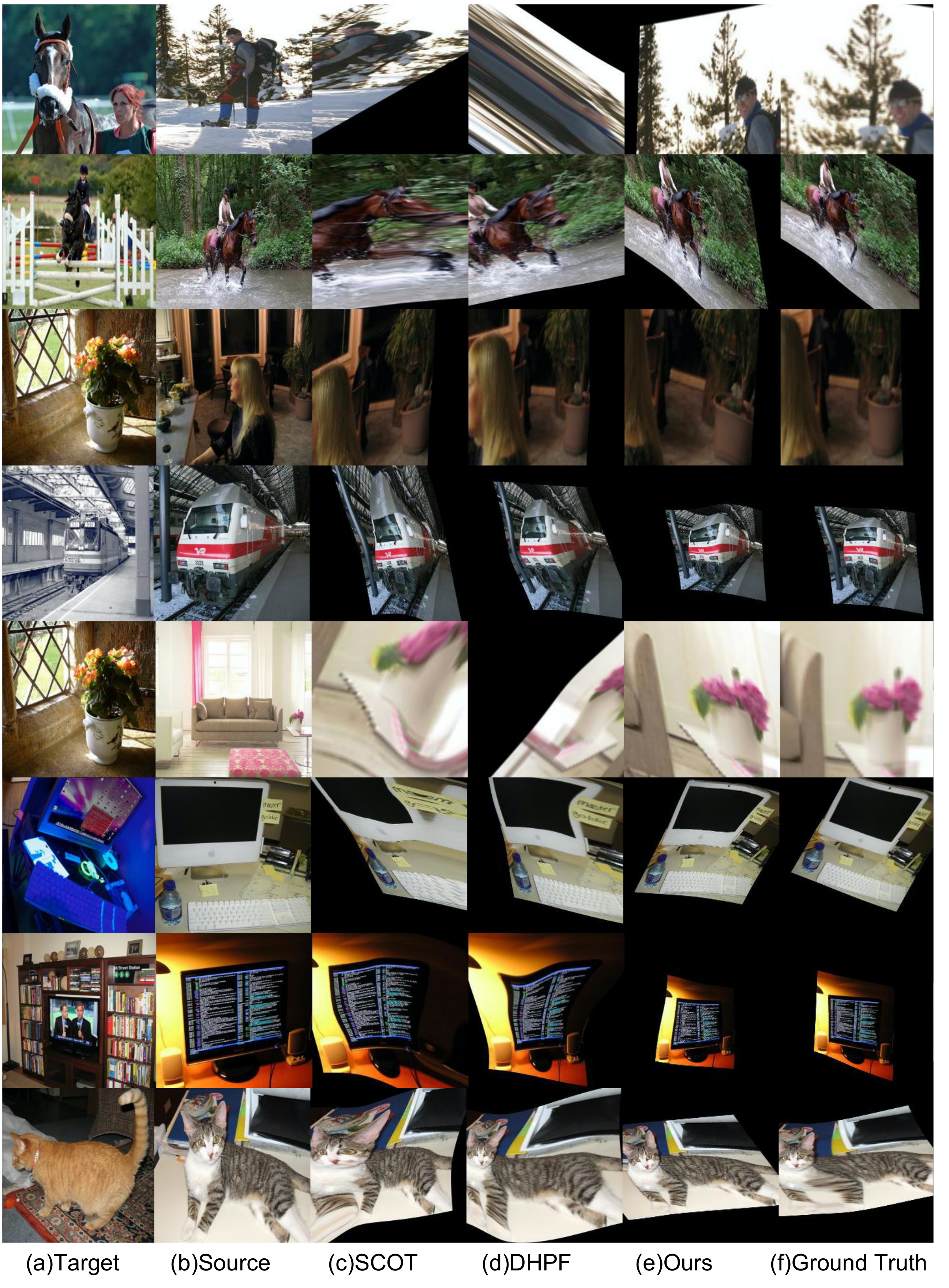}
	\caption{Warped images by thin-plate splines with the predicted key point pairs on SPair-71k dataset  \cite{spair} compared with SCOT \cite{liu2020semantic} and DHPF \cite{min2020learning}}
	\label{Fig:warp_2}
\end{figure*}

\end{document}